\documentclass[10pt,letterpaper]{article}

\usepackage{cogsci}
\usepackage{pslatex}
\usepackage{apacite}

\usepackage{latexsym,amssymb,lastpage}
\usepackage{graphicx,amsfonts}
\usepackage{times,mathptmx,bm,amsmath}
\usepackage{dcolumn}
\usepackage{epsfig}

\title{Exemplar Dynamics Models of\\ the Stability of Phonological Categories}
 
\author{{\large \bf P. F. Tupper (pft3@sfu.ca)} \\
  Department of Mathematics, Simon Fraser University, 8888 University Drive \\
  Burnaby, BC V5A 1S6 Canada}

\begin{document}

\maketitle

\begin{abstract}
We develop a model for the stability and maintenance of phonological categories.  Examples of phonological categories are vowel sounds such as \emph{i} and \emph{e}.  We model such categories as consisting of collections of labeled exemplars that language users store in their memory.  Each exemplar is a detailed memory of an instance of the linguistic entity in question.  Starting from an exemplar-level model we derive integro-differential equations for the long-term evolution of the density of exemplars in different portions of phonetic space. Using these latter equations we investigate under what conditions two phonological categories merge or not. Our main conclusion is that for the preservation of distinct phonological categories, it is necessary that anomalous speech tokens of a given category are  discarded, and not merely stored in memory as an exemplar of another category.

\end{abstract}

\section{Introduction}

%
%

Our goal is a model of how phonological categories, like those of the vowel sounds \emph{i} and \emph{e},  are formed and are maintained,  and what laws govern their dynamics.
For this paper we consider a single important aspect of these systems: whether or not two  categories merge. Consider the pair of words \emph{cot} and \emph{caught}. In many dialects of English these two words are homophonous, i.e.\ they are pronounced the same way. At some point in the past they were pronounced distinctly, but over time the words became indistinguishable when spoken out of context. We say that the phonological categories corresponding to the vowels in these words merged in this context.


One theory for why phonological categories merge in some contexts but not in others is that sounds merge when there is no harm done to communication by them having merged. For example, we may suppose that the words \emph{cot} and \emph{caught} being pronounced the same way is harmless for communication, since which word is intended can almost always be determined from context. In contrast, a merger between the pronunciations of the words \emph{can} and \emph{can't} could lead to serious miscommunications. However, to actually explain why phonological categories merge in some contexts and not in others, what is needed is an actual mechanism for the merger or non-merger of phonological categories.
We demonstrate a set of features of a model of phonological categories that are sufficient to explain this phenomenon.

%

To model the  dynamics  of phonological categories we need a model of how such categories are stored in the brain.
We use \emph{exemplar models} \cite{nosofsky2,johnson} in which categories consist of collections of labeled \emph{exemplars}. Exemplars are instances of stimuli from the given category which are stored in detailed representation in the subject's memory.
 To model the merger or non-merger of phonological categories we use \emph{exemplar dynamics} \cite{pierrehumbert_exemplar}, in which the long term behaviour of categories is modelled as emerging from   the bulk behaviour of many exemplars in language-users' memory.

We consider a particular class of exemplar dynamics models building on those of Pierrehumbert (2001), Wedel (2006), Ettlinger (2007), and Blevins and Wedel (2009).
In our model every language-user has a small number of linguistic categories.  Each category consists of exemplars which are detailed memories of instances of the category in question. 
For the sake of simplicity, in our model each exemplar will be described by only 1 or 2 scalars indicating  phonetic variables. For example, if we are modelling vowel categories, we may imagine exemplars that consist of the two phonetic variables F1 and F2. 
Each exemplar has a weight that decays with time, representing the strength of the exemplar in memory. When a language-user, the speaker, wishes to utter a representative of a category, an exemplar is selected from that  category randomly according to the exemplars' weights. The new exemplar's weight starts off at some fixed large value. The new exemplar's position is close to the original exemplar but is biased towards less extreme values (known as lenition) and may be biased towards the category mean (known as entrenchent \cite{pierrehumbert_exemplar}). Random noise is also added to the position of the new exemplar.  This new exemplar is heard by another language-user, the listener, and this listener classifies the exemplar into one of their categories.

Exemplar dynamics models  have some features that make them difficult to work with computationally. On the one hand, the computational expense of simulating an exemplar model grows with the number of exemplars in the system. Phenomena that only emerge in exemplar dynamics with millions or billions of exemplars in the system may be missed in simulations of smaller systems. On the other hand, exemplar models involve randomness, through both the selection of exemplars to be reproduced, and in the noise added in the production of new exemplars. This means that the results of simulations of exemplar models are random. In order to determine if a phenomenon in an exemplar simulation is a rare aberration or quite typical, it is necessary to do several exemplar simulations with the same parameters, adding to the cost of studying the system computationally. This latter effect becomes more pronounced for small numbers of exemplars.

To overcome these difficulties we study exemplar models in the limit where systems contain infinitely many exemplars. In this limit, rather than the location and weight of every exemplar in phonetic space, we consider fields which describe the density of exemplars of different types in different locations in phonetic space. We derive integro-differential equations for these exemplar density fields, which we refer to as \emph{field models} corresponding to the exemplar model.    We study the field model in order to infer properties of the exemplar model in the limit of very many exemplars.

Through simulations of exemplar models and the limiting field models, we show in this paper that it is the procedure by which a language-user classifies heard exemplars that determines whether or not two categories merge. If the language user always classifies heard exemplars accurately, no matter how they are produced, then the phonological categories will merge. This is believed to be what happens in situations where the category can be inferred from context. However, if there is the potential for exemplars to be misclassified, then the categories will remain distinct, under some additional modelling assumptions.
We find that an important aspect of our model for the categories to behave realistically in the non-merger case
  is that exemplars that are misclassified are discarded, rather than stored in memory.

In order to have tractable models of the phenomena of interest, in all that follows we have made a simplifying assumption: rather than modelling each individual in a population of language-users with their own exemplars, we pool all the exemplars in the population.  This will not make a significant difference to our results if each individual in the population has the same categories with the same statistical characteristics of exemplars within their categories. 
Accordingly, our models and simulations should not be thought of as describing a single language-user's  phonological categories changing over their lifetime, but of the change of the phonological categories of an entire population of language-users over possibly many generations.

%
%
%

\section{One Category} \label{sec:onecategory}

We first describe the exemplar-level model with exemplars possessing a  single phonetic variable and belonging to a single category.  We then describe the corresponding field model.

\subsection{One Category Exemplar Model} \label{subsec:onecat_exemp}

At any point in time the state of the system consists of the positions of $n$ exemplars $y_1, \ldots, y_n$, and their respective weights $w_1, \ldots, w_n \geq 0$.
Initially there are $n(0)$ exemplars with fixed positions and weights. The number of exemplars at time $t$, $n(t)$ is non-decreasing, and increases  through the creation of new exemplars.
 Once an exemplar is  created, its position does not change and it is never removed from the system, though its weight decreases.  New exemplars are introduced to the system
 at a  constant rate $\nu$. The weight of new exemplars is $W_0$. The weight of all exemplars decay with rate $\lambda$, so that exemplars that have been around for long enough cease to have significant effect on the system.  The position of new exemplars is determined through two steps. First, one of the existing exemplars is selected with probability proportional to its weight. We imagine that this exemplar is being spoken by a speaker.  Let $z$ be the position of this exemplar. A copy of this exemplar is produced with a position moved towards the centre of the phonetic space (lenition) and also towards the mean of the exemplar positions (entrenchment).  Let $\bar{y}$ be  the weighted mean position of all the exemplars:
\[
\bar{y}= \frac{\sum_{i=1}^n w_i y_i  }{\sum_{i=1}^n w_i}.
\]
 Then the new exemplar is generated at position
\begin{eqnarray*}
y & = & (1- \beta) z + \alpha (\bar{y}-z) + \sigma \eta
\end{eqnarray*}
where $\eta$ is a standard Gaussian random variable, independent of all other variables, and $\alpha$, $\beta$, $\sigma$ are parameters representing respectively the rate of pull to zero, the rate of pull to $\bar{y}$, and the magnitude of errors in production/perception.

\begin{figure}[ht]
\begin{center}
\includegraphics[width=8.5cm]{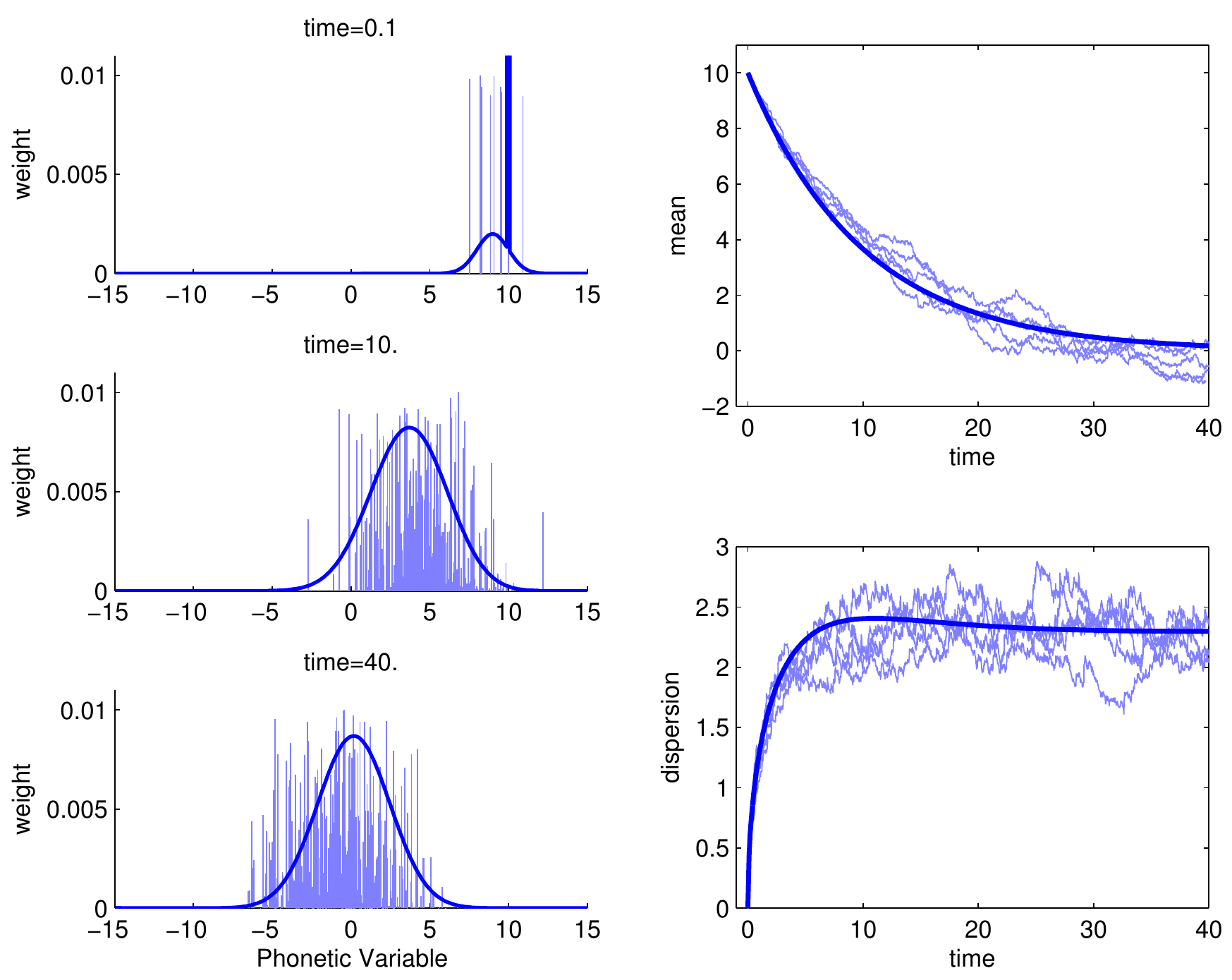}
\end{center}
\caption{  \label{fig:one_category_exemplars} In light blue, results for the exemplar model in one dimension for  one category. $\nu=100$, $\lambda=1$, $W_0=1/\nu$, $\alpha=0$, $\beta=.1$.  On the left we show the exemplars in the system at three different points in time, for a single realization. On the right we show the mean and dispersion of the category as a function of time for five different realizations.  In dark blue, corresponding results for the field model.}
\end{figure}

In Figure~\ref{fig:one_category_exemplars} we show the results of a simulation of this model. The light blue lines indicate results from the exemplar model, and the dark blue lines indicate results from the field model we will derive in the next subsection. On the left we show the location and weight of all the exemplars in the system at three points in time. On the top right we show how the  mean of the exemplars' phonetic variable varies with time. We have plotted the results for five independent runs of the simulation. The mean converges to nearby 0 in each simulation.
On the bottom right, again for five different simulations, we show how
 the \emph{dispersion} of the category varies with time, where dispersion defined by
\[
\sqrt{ \frac{\sum_{i=1}^n w_i (y_i-\bar{y})^2  }{\sum_{i=1}^n w_i}},
\]
is the weighted standard deviation of the phonetic variable of the exemplars about their mean. Dispersion is a measure of the width of a category in phonetic space.

\subsection{One Category Field Model} \label{subsec:onecat_nonloc}

Recall that the state of the exemplar model at time $t$ consists of $n(t)$ exemplars with positions $y_i$ and weights $w_i$.  We define the exemplar density field to be
\[
\rho(t,y) = \sum_{i=1}^n \delta(y-y_i)w_i(t),
\]
where $\delta$ is the $\delta$-distribution. Roughly, the field consists of of infinitely narrow peaks at locations $y_i$ with weights $w_i$.
We derive equations for the limit of this field as rate of exemplar production $\nu$ goes to infinity and initial exemplar weight $W_0 \rightarrow 0$.   

Recall that new exemplars are produced at rate $\nu$ and that  the weight of each exemplar decays at rate $\lambda$.  
If each exemplar is created with weight $W_0$ then we must have $W_0 \nu=\lambda W$ at steady state where $W$ is the total weight in steady state.  We fix $\lambda$ and let $W_0$ go to zero while $\nu$ goes to infinity.
$\nu$ and $W_0$ should be inversely related to guarantee a fixed $W$ limit, so we let $W_0=\mu/\nu$, where $\mu$ is a constant. In this limit the equations describing how $\rho(t,y)$ changes with time are  
\begin{equation}\label{eq:onecat_nonloc}
\frac{\partial \rho(y)}{\partial t} = - \lambda \rho(y) +  \mu \frac{\int f(y- z + \beta z - \alpha (\bar{y}-z) ) \rho(z) dz }{\int \rho(z) dz}
\end{equation}
 where $f$ is the density of Gaussian with mean zero and variance $\sigma^2$. This derivation will be given in a separate paper.
 There we also show that the equilibrium value of $\rho$, which the system will converge to, is a Gaussian given by
%
%
%
\[
\rho(y) = \frac{\mu}{\lambda} \frac{1}{\sqrt{2 \pi s^2}} e^{ - y^2/ 2 s^2} 
\]
where the dispersion $s$ is given by
\[
s^2 : = \frac{ \int (y-\bar{y})^2 \rho(y) \, dy}{\int \rho(y) \, dy}  = \frac{\sigma^2}{ 1 - (1 - (\alpha+\beta)) ^2}
\]
Note that $\alpha$ and $\beta$ do not have independent effects on the equilibrium.
In Figure~\ref{fig:one_category_exemplars} the dark blue lines show the results of simulation of this field model, which we see captures many of the features of the exemplar model.

\section{Multiple categories}

In this section we generalize our models  to describe the interactions of multiple phonological categories. Production and decay of exemplars proceeds as before, but after a new exemplar is produced it is assigned to one of the existing categories according to a classification procedure. We develop the models for the case of two interacting categories here; the extension to more than two is straightforward.


\subsection{Multiple Category Exemplar Model} \label{subsec:marmot1}

Each exemplar now belongs to one of two categories, A and B. Exemplars are produced from each category with rates $\nu_A$, $\nu_B$ respectively. As before, all exemplar are produced with weight $W_0$ and decay with rate $\lambda$.

To describe how categorization works, suppose a new exemplar is produced from category A.  An existing exemplar in category A is selected randomly with probability proportional to exemplar weight.  If $z$ is the position of the existing exemplar, a new exemplar is  produced at 
\[
y= (1-\beta) z + \alpha (\bar{y}_A - z)  + \sigma \eta
\]
where $\bar{y}_A$ is the weighted mean of the positions of the exemplars in category $A$.

The new exemplar has to be classified as either belonging to category A, classified as belonging to category B, or discarded. We consider three different ways that this may occur, which we call  \textit{ \textbf{Categorization Regimes}}. 

\vspace{-.2cm}
\begin{enumerate} \setlength{\itemsep}{-.1cm}
\item {\emph{No Competition:}} Exemplars are always classified into the category they were produced from.
\item {\emph{Pure Competition:}} Categories compete to claim the new exemplar as their own, without information about the category it came from. The winner accepts the exemplar in its category.
\item {\emph{Competition with Discards:}} Categories compete to claim the new exemplar as their own, without information about the category it came from. If the exemplar is claimed by the category it came from, then it is stored in that category. Otherwise it is discarded.
\end{enumerate}
\vspace{-.2cm}

In the first, the \emph{No Competition} regime, the exemplar is assigned to the category that it was produced from. Linguistically, this means that the listener is able to figure out from context what category was intended, regardless of the actual sound produced by the speaker. In the non-competitive case there is no interaction between exemplars of different varieties. So the model consists of the exemplars for category A and the exemplars for category B both evolving independently of one another.

 In the second, the \emph{Pure Competition} regime, the listener has to decide which category to classify the exemplar in based on the phonetic variables of the exemplar: the categories ``compete" to acquire the exemplar.
The idea is that the exemplar is more likely to be classified in category $A$ if there are more exemplars of category $A$ with strong weight that are close to the new exemplar.  
We assume the category of its parent exemplar does not figure in the assignment. This is the mechanism for prevention of the merger of categories proposed in Blevins and Wedel (2009).

In the third, the \emph{Competition with Discards} regime, the listener decides how to classify the exemplar exactly as in the purely competitive case. But the listener is aware of what the intended category is, and if the perceived category does not match the intended category, the exemplar is discarded. This is a mechanism for prevention of the merger of categories proposed by  Labov (1994, pp. 586--588); see also \cite[Ch.\ 5]{silverman}.

%
%
%
%
%
%

In the two competitive regimes we need a model of how the categories compete for an exemplar.
In order for the classification of a new exemplar to be influenced by the weight and number of nearby exemplars  of both categories, we define local smoothed category density fields:
\begin{equation} \label{eqn:hyrax1}
S_A(y)  =  \sum_{i=1}^{n_A} \frac{1}{2} \kappa w_{A,i} e^{-\kappa |y-y_{A,i}|}, \ \ \ \
 S_B(y)  =  \sum_{i=1}^{n_B} \frac{1}{2} \kappa  w_{B,i}  e^{-\kappa |y-y_{B,i}|}.
\end{equation}
The factor of $\frac{1}{2}\kappa$ guarantees that $\int_y S_A(y) dy = \sum_i w_{A,i}$ and likewise for category $B$. As $\kappa \rightarrow 0$ the smoothed fields are uniform in space.  As $\kappa \rightarrow \infty$ there is no smoothing and $S_A(y) = \rho_A(y)$, $S_B(y)=\rho_B(y)$. In the simulations that follow we set $\kappa=10$, so that a new exemplar is classified mostly depending on the category of other exemplars within about $1/\kappa=0.1$ of it.
We define the  probability of the new exemplar generated at point $y$ being assigned to category $A$ or $B$ as  
\begin{equation*}
f_A(y) =  \frac{S_A(y)^p}{S_A(y)^p +S_B(y)^p}, \ \ \ \ \ f_B(y) =  \frac{S_B(y)^p}{S_A(y)^p +S_B(y)^p},
\end{equation*}
respectively,
where $p \geq 0$ is a fixed selection parameter. For $p=1$ the exemplar is assigned to a new category in direct proportion to the magnitude of the smoothed field at $y$. For $p\rightarrow \infty$ limit the exemplar is assigned to whichever category has greater density.

\subsection{Multiple Category Field Model} \label{subsec:marmot2}
 
 Taking the same limit as we do in the single-category case we can derive equations for the exemplar density fields.
 
 As in the single category case, we define the exemplar density fields $\rho_A$ and $\rho_B$. We also define field analogues of the smoothed density, as well as the selection functions $f_A$ and $f_B$, which also depend on a selection parameter $p$. We define the production terms $P_A$ and $P_B$, which describes the rate of production of new exemplars of type $A$ and $B$ respectively at location $y$. 
 For each categorization regime we can derive a different set of coupled equations for $\rho_A$ and $\rho_B$.
 For example, for the Categorization with Discards regime the equations are
 \begin{equation*} \label{eqn:compdisc}
 \begin{aligned}
\frac{\partial \rho_A(y)}{\partial t} & =  - \lambda \rho_A(y) +  f_A(y) 
P_A(t,y), \\
\frac{\partial \rho_B(y)}{\partial t} & =  - \lambda \rho_B(y) +  f_B(y)  P_B(t,y).
\end{aligned}
\end{equation*}

\subsection{Numerical Simulations of Multiple Categories} \label{subsec:numer}

Here we explore  the behaviour of the exemplar model and the  field model with two categories under the different categorization regimes. 
In all of the following simulations we choose $\alpha=0$, meaning that there is no bias in production towards the category mean. (Having $\alpha>0$ does not qualitatively change the results.) In each case we also set  $\beta= 0.1$, $\lambda=1$, $\kappa=10$, and $W_0=1/\nu$, leading to $\mu=1$. We vary the categorization regime and the selection parameter $p$. 

\begin{figure}[ht]
\begin{center}
\includegraphics[width=4.12cm]{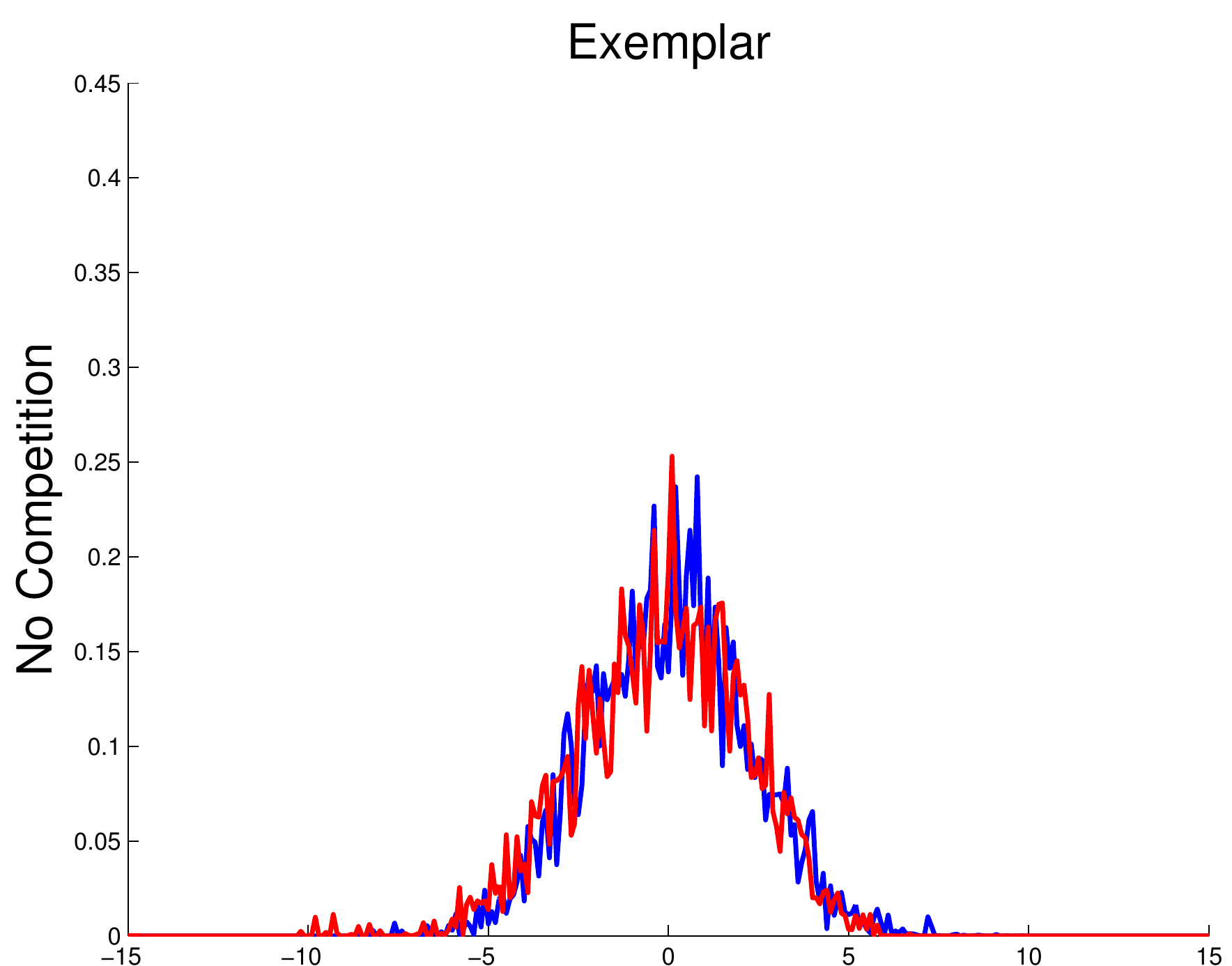}
\includegraphics[width=3.88cm]{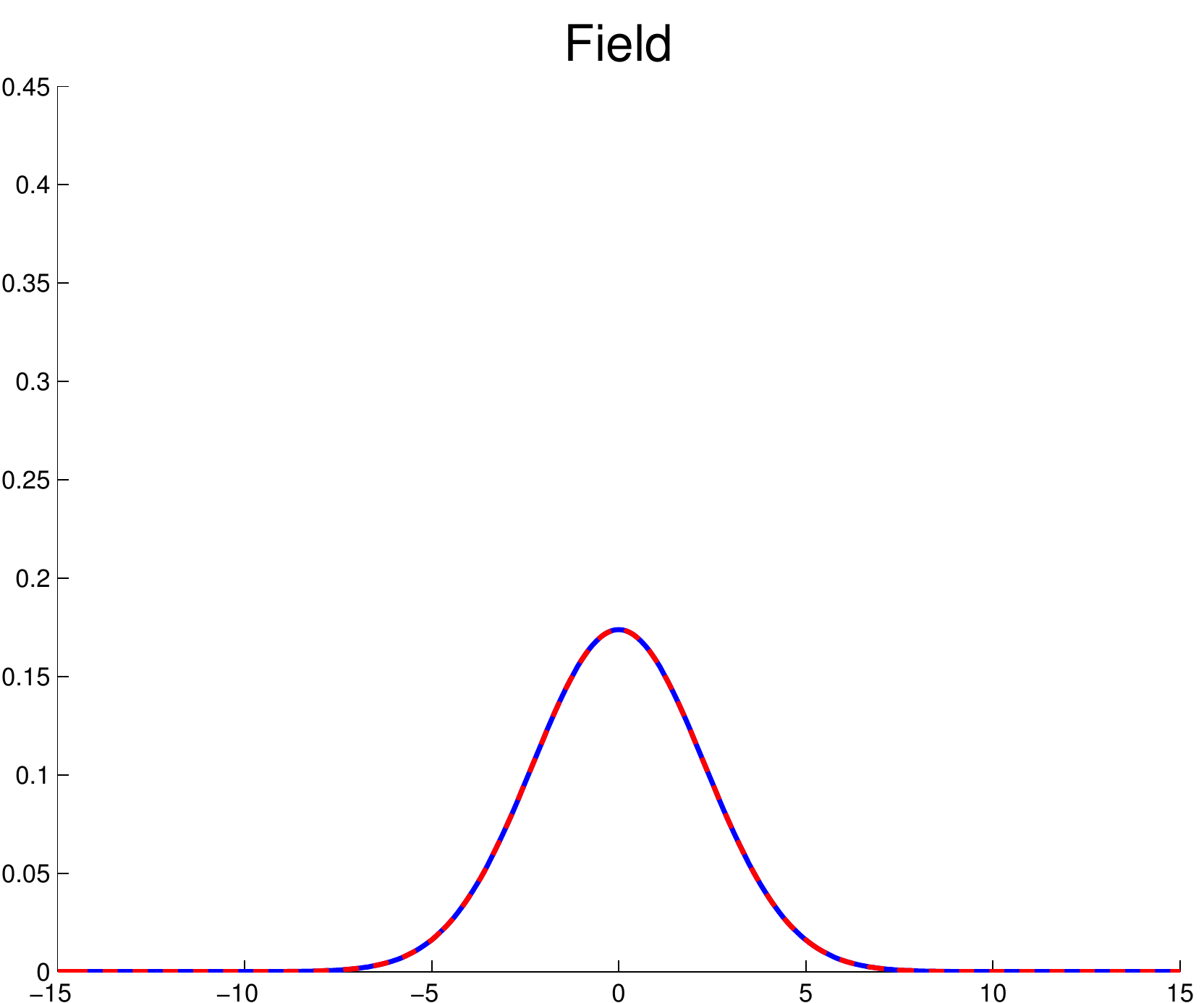}
\includegraphics[width=4.12cm]{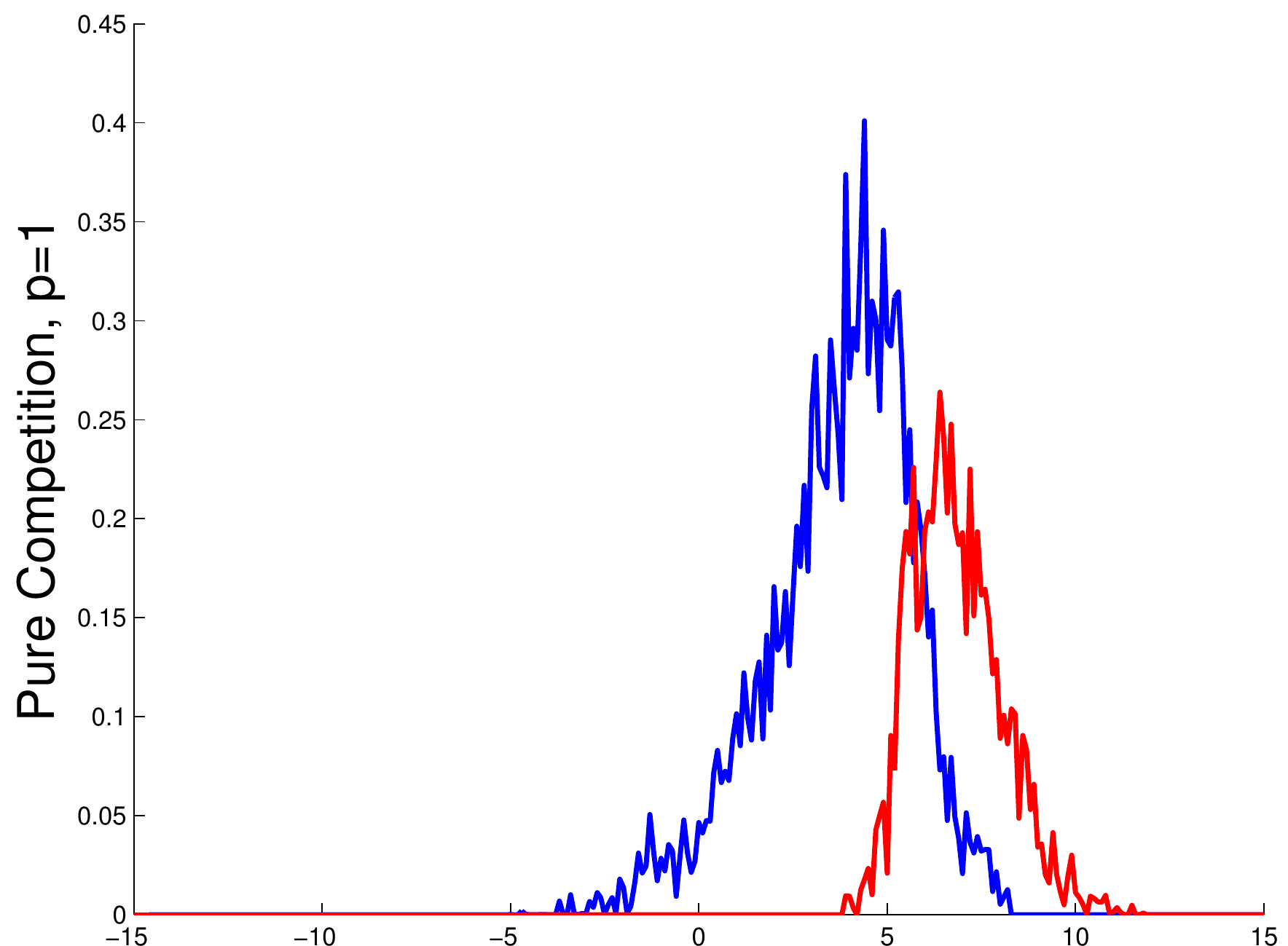}
\includegraphics[width=3.88cm]{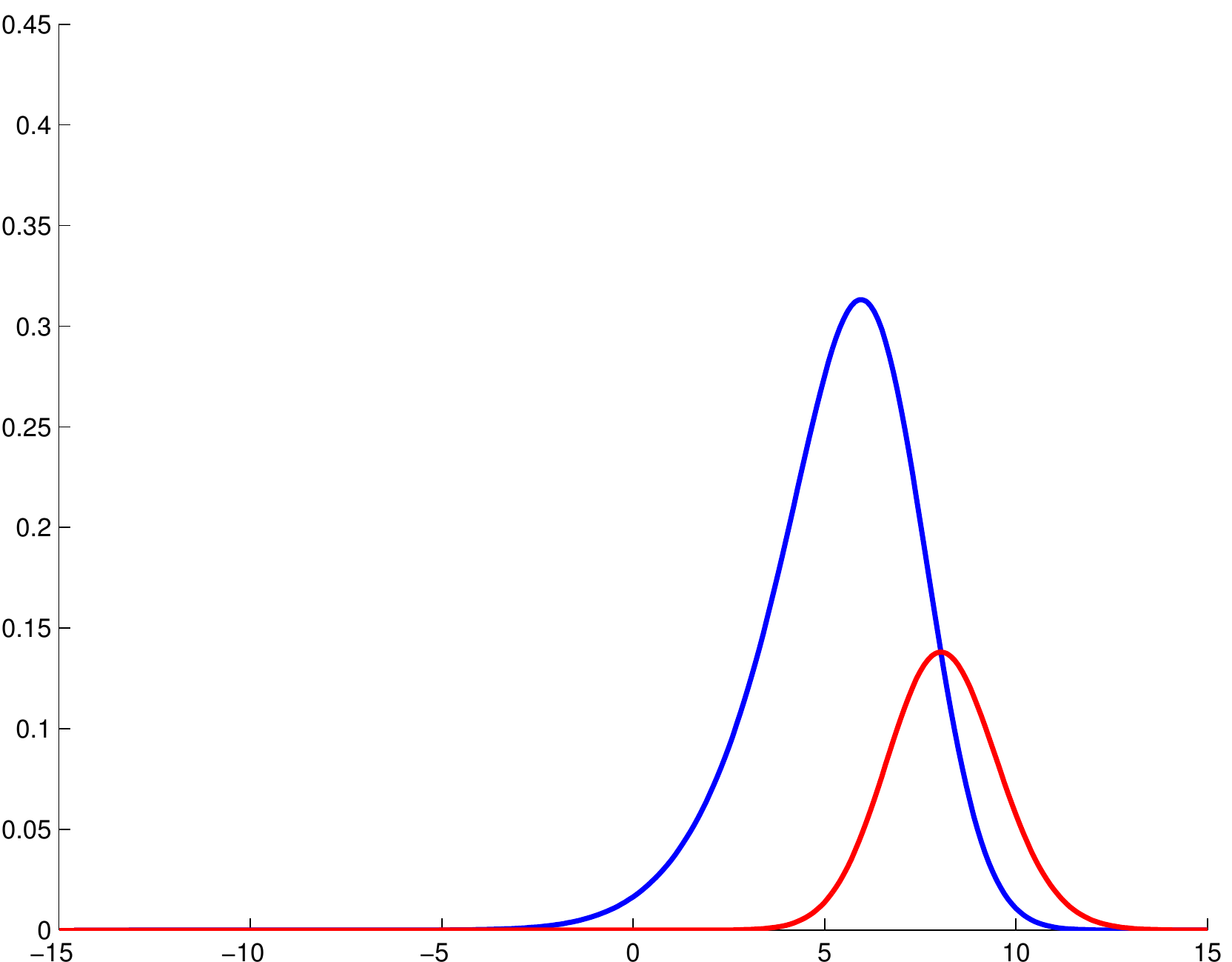}
\includegraphics[width=4.12cm]{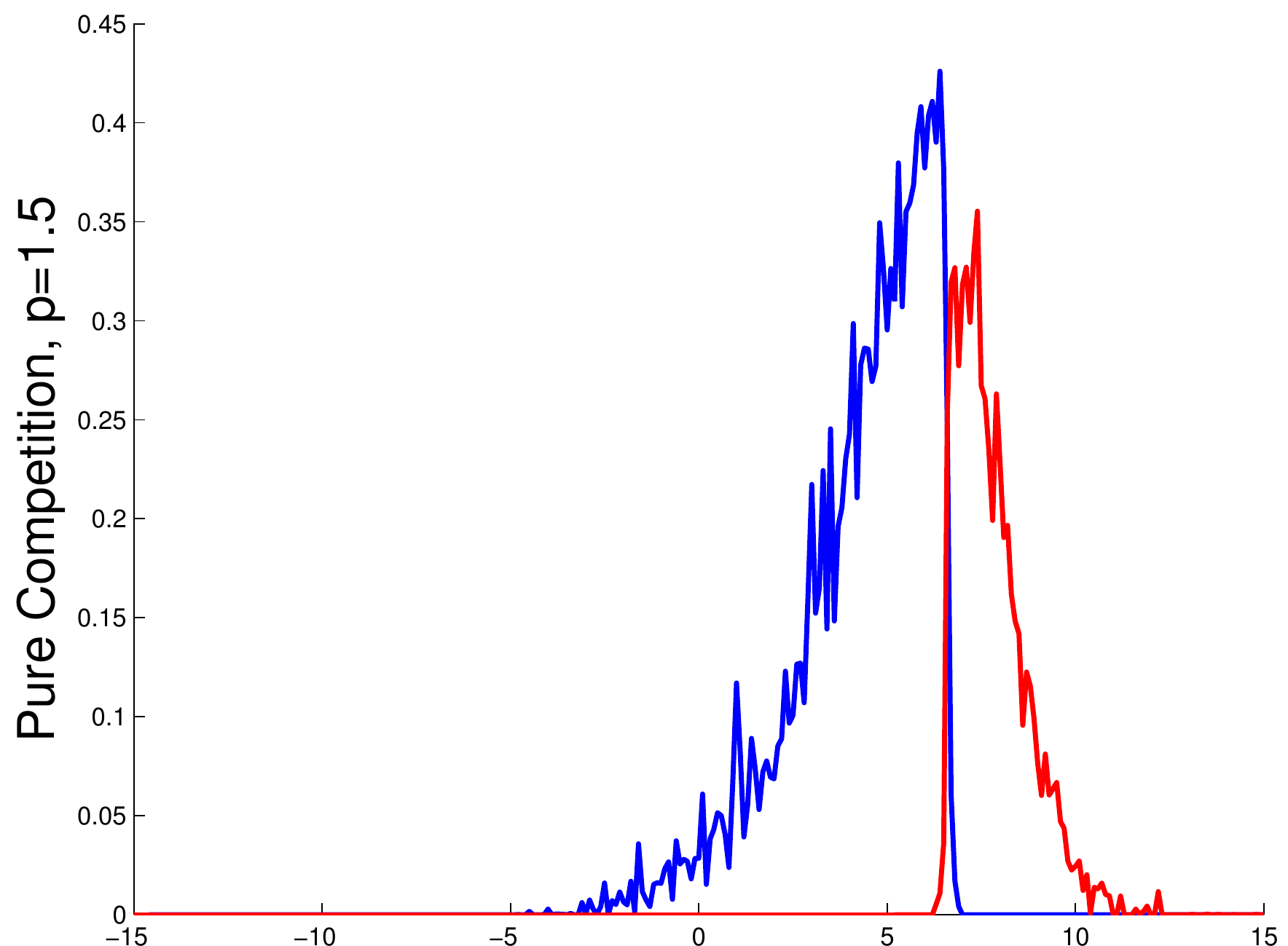}
\includegraphics[width=3.88cm]{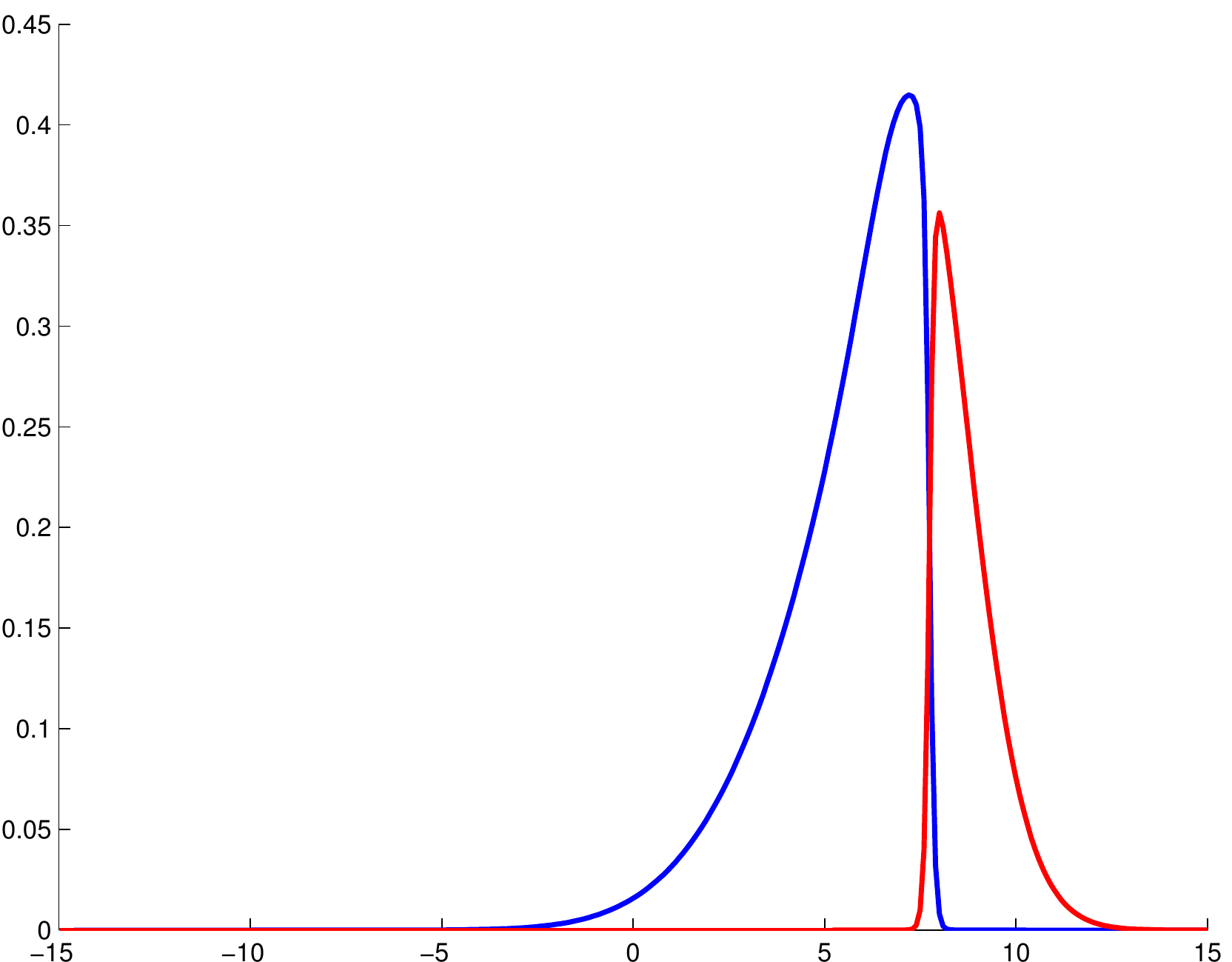}
\includegraphics[width=4.12cm]{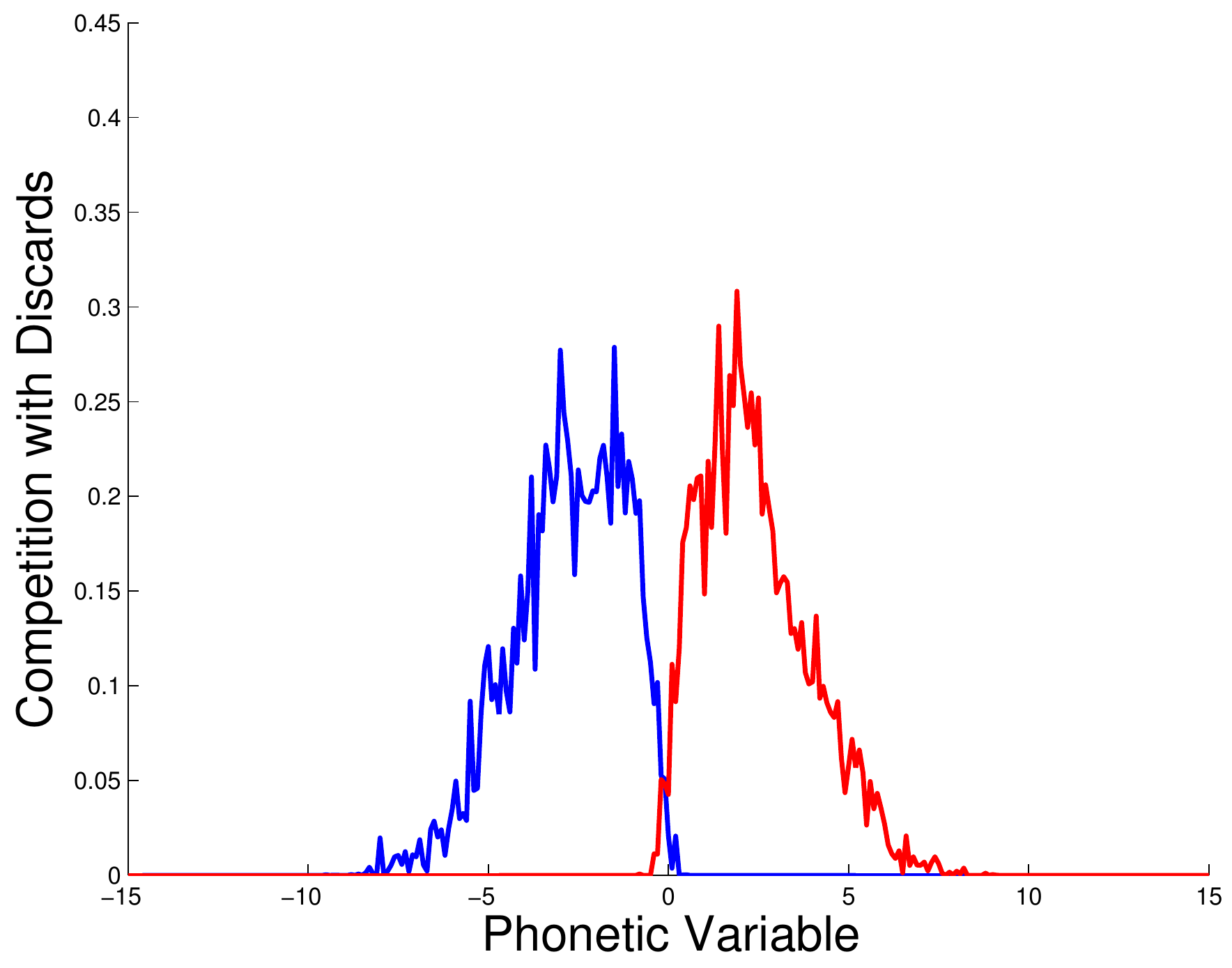}
\includegraphics[width=3.88cm]{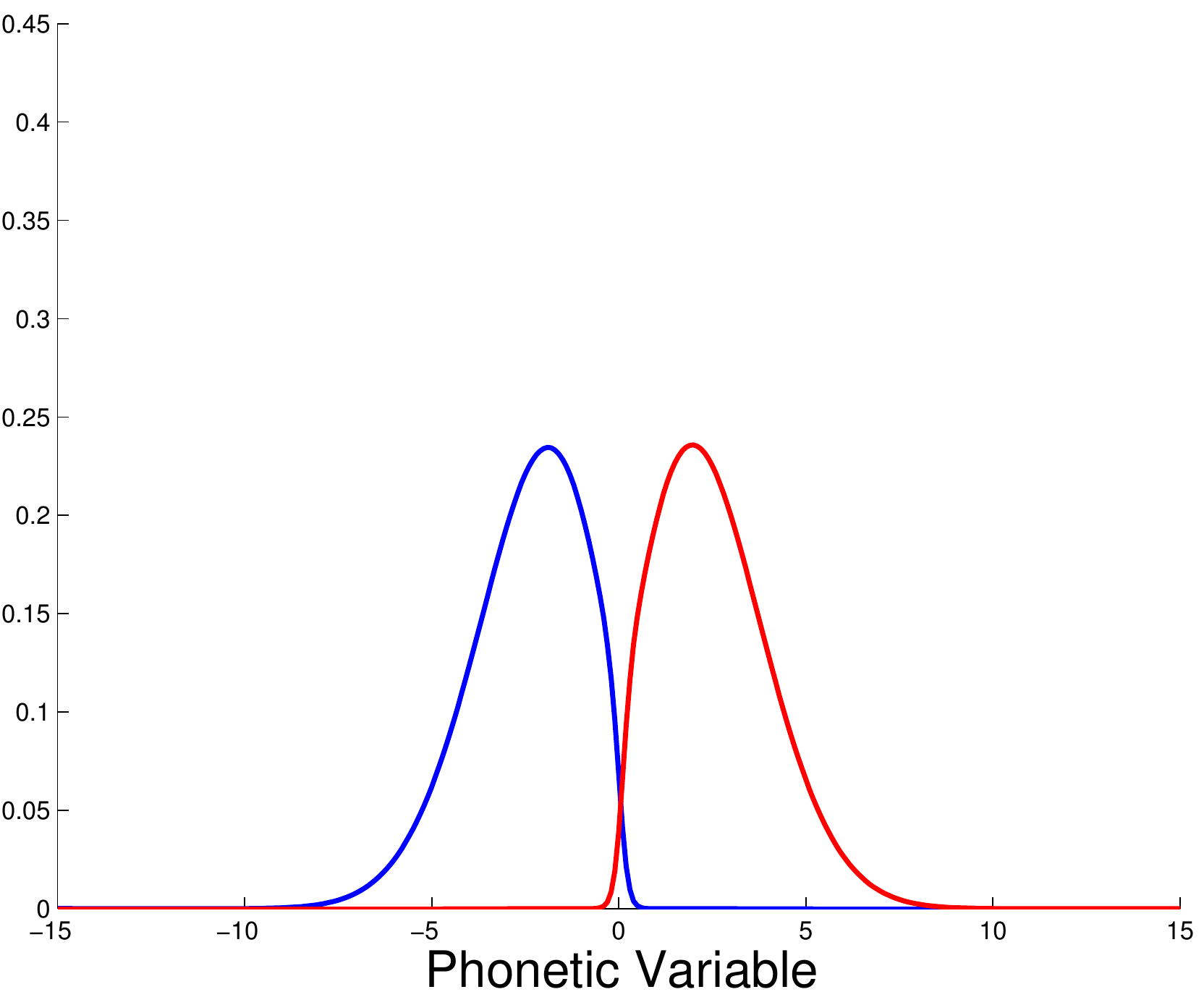}
\end{center}
\caption{ \label{fig:blah} Exemplar density fields $\rho_A$ (blue) and $\rho_B$ (red) for the exemplar model (left) and the field model (right) at $t=100$. Four different categorization regimes/choice of selection parameter $p$ are shown. }
\end{figure}

\begin{figure}[ht]
\begin{center}
\includegraphics[width=8.5cm]{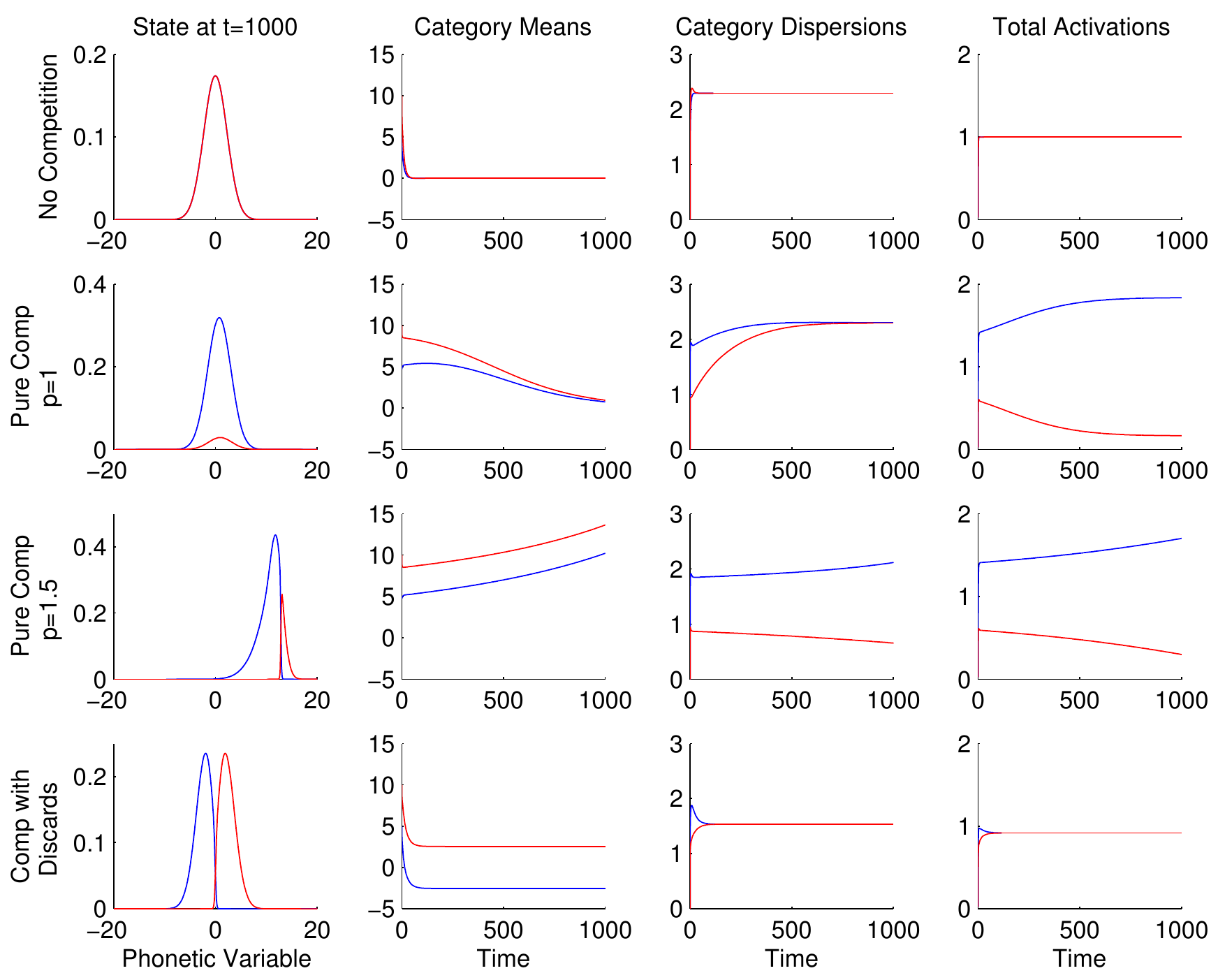}
\end{center}
\caption{\label{fig:long} Each row shows results for the field simulation with a different categorization regime.
The first column shows the exemplar density  $\rho$ as a function of $y$ at time $t=1000$.
}
\end{figure}

 For each regime we present results for the models over 100 time units. For the exemplar model we use $\nu=10^3$ and  initialize the system with $50$ exemplars of category $A$ at $y=5$ and $50$ exemplars of category $B$ at $y=10$. The field model is initialized with a matching initial condition. The results of these simulation are presented in Figure~\ref{fig:blah}.  We show a plot of $\rho_A$ and $\rho_B$ at time $t=100$, with the exemplar model on the left and the corresponding field model on the right.
  We also present in Figure~\ref{fig:long}  results of a longer simulation  (over 1000 times units) of the field model alone. In this latter figure we show $\rho_A$ and $\rho_B$ at time $t=1000$ in the first column, and in the remaining three columns we plot category means, category dispersions, and total category activation (sum of weights) versus time. We discuss each of the different choices of categorization regime and selection parameter $p$ in turn.

\newpage

\emph{Noncompetitive:}
Using the noncompetitive classification regime the system is just two uncoupled versions of the single category case. 
Both $\rho_A$ and $\rho_B$  rapidly converge to the steady state Gaussian of the single category case, as can be seen in the top row of Figure~\ref{fig:blah} and  in the top row of Figure~\ref{fig:long}.

\emph{Purely Competitive: $p=1$.} 
We show the results for this regime in the second rows of Figure~\ref{fig:blah} and Figure~\ref{fig:long}. 
The plots demonstrate that even though the two categories start off separated, they slowly merge over the course of the simulation.  This impression is confirmed by a longer simulation of the field model which is shown in the second row of Figure~\ref{fig:long}. At $t=1000$ the categories have nearly the same means, and the means appear to be converging to zero. 
In a separate paper we demonstrate that this phenomenon is an artefact of the choice $p=1$.
We conclude that, even though it may be a realistic model of categorization in some contexts,  the purely competitive regime with selection parameter $p=1$ cannot account for the existence of  stable distinct categories in phonetic systems.

\emph{Purely Competitive: $p=1.5$.}
The argument just made for why the pure competitive regime leads to category merger does not hold when the selection parameter $p>1$. Indeed, with this choice, the more active category in a region of phonetic space draws in more than its share of new exemplars. To test this idea, we perform the same simulation as we did for $p=1$ for $p=1.5$. The results are shown in the third rows of  Figure~\ref{fig:blah} and Figure~\ref{fig:long}.
 
 As predicted, the categories do not merge and appear to remain sharply distinct for all time. However, there is another anomaly: the system does not appear to settle into an equilibrium that is symmetric about $y=0$.
In Figure~\ref{fig:blah} it seems that the categories have a boundary near $y=8$ at time $t=100$. 
 Looking at the longer simulation of the field model in  the third row of Figure~\ref{fig:long} we see that in fact the categories are slowly moving to the right.  We conjecture that both categories continue to move to the right together indefinitely. 

 
%


Again, the Pure Competition regime with $p=1.5$ may be a realistic model in some circumstances. However, it is unlikely to be a very general mechanism for the stability of phonological categories, since even though the categories do not merge, they are not stably located within phonetic space.

\emph{Competition with Discards. $p=1$.} Finally we consider Competition with Discards categorization regime. Choosing selection parameter $p=1$, we plot the final state in the bottom row of Figure~\ref{fig:blah}. Here the system approaches a steady state where the two categories occupy distinct portions of phonetic space they are symmetric about $y=0$.  This is confirmed by the longer simulation of the field model shown in the fourth row of Figure~\ref{fig:long}.

This categorization regime appears to have the best hope of capturing the features of phonological categories in a linguistic context, and so we explore its behaviour further for two phonetic variables and more than two categories.

%

%
%
%
%
%
%
%
%

 \section{Two Dimensional Model and Simulation}

The models we have presented so far extend immediately to more than one phonetic  dimension.
We show the results of  two-dimensional simulations  of both an exemplar and its corresponding field model. Both simulations were done with the 5 categories and the Competition with Discards regime for categorization. The parameters in the model are set as follows for both simulations: $\lambda=1$, $\mu=1$, $\sigma=1$, $\beta=0.1$, $p=1$, $\kappa=10$.

\begin{figure}[ht]
\begin{center}
\includegraphics[width=4cm]{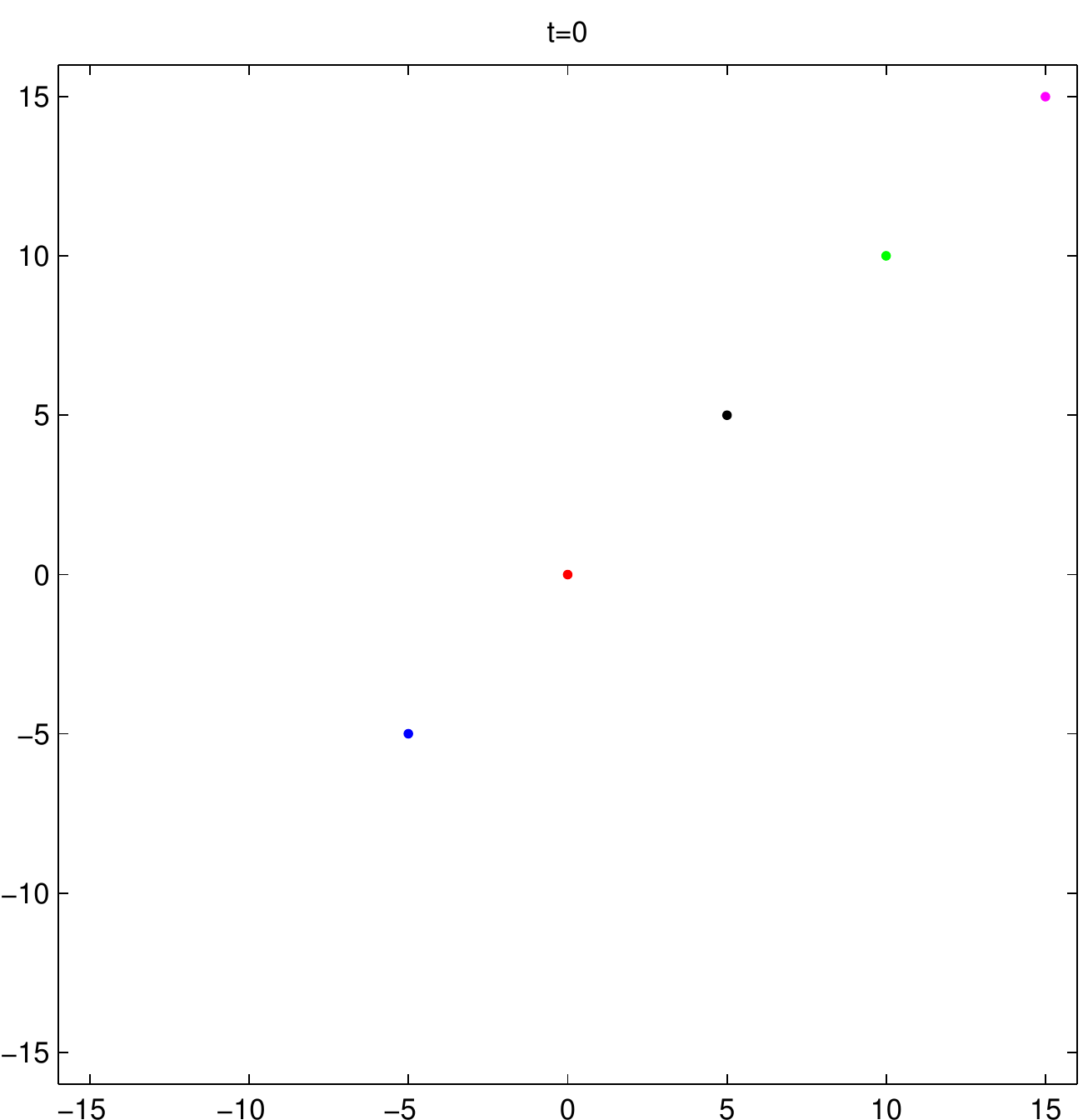}
\includegraphics[width=4cm]{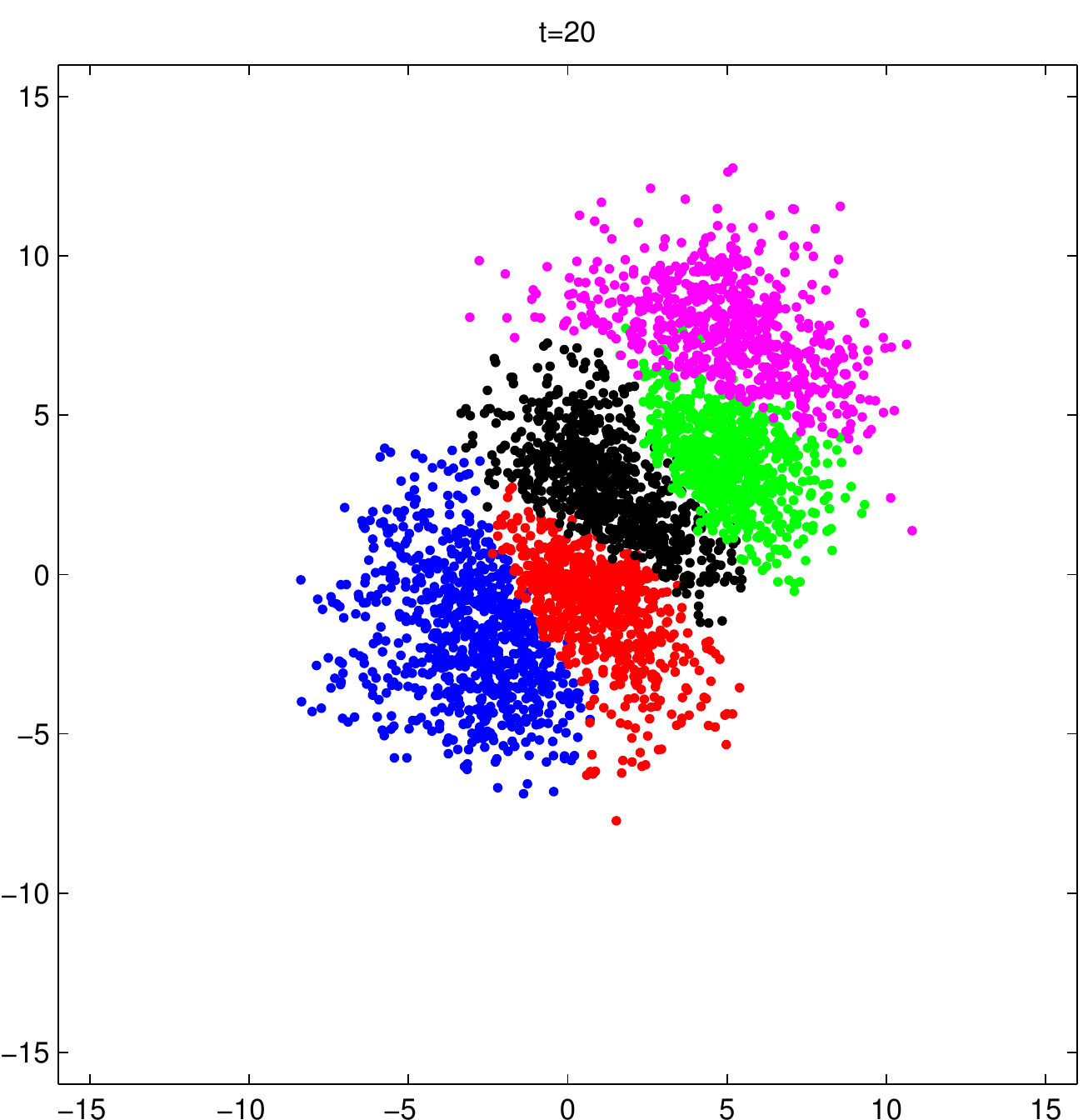}
\includegraphics[width=4cm]{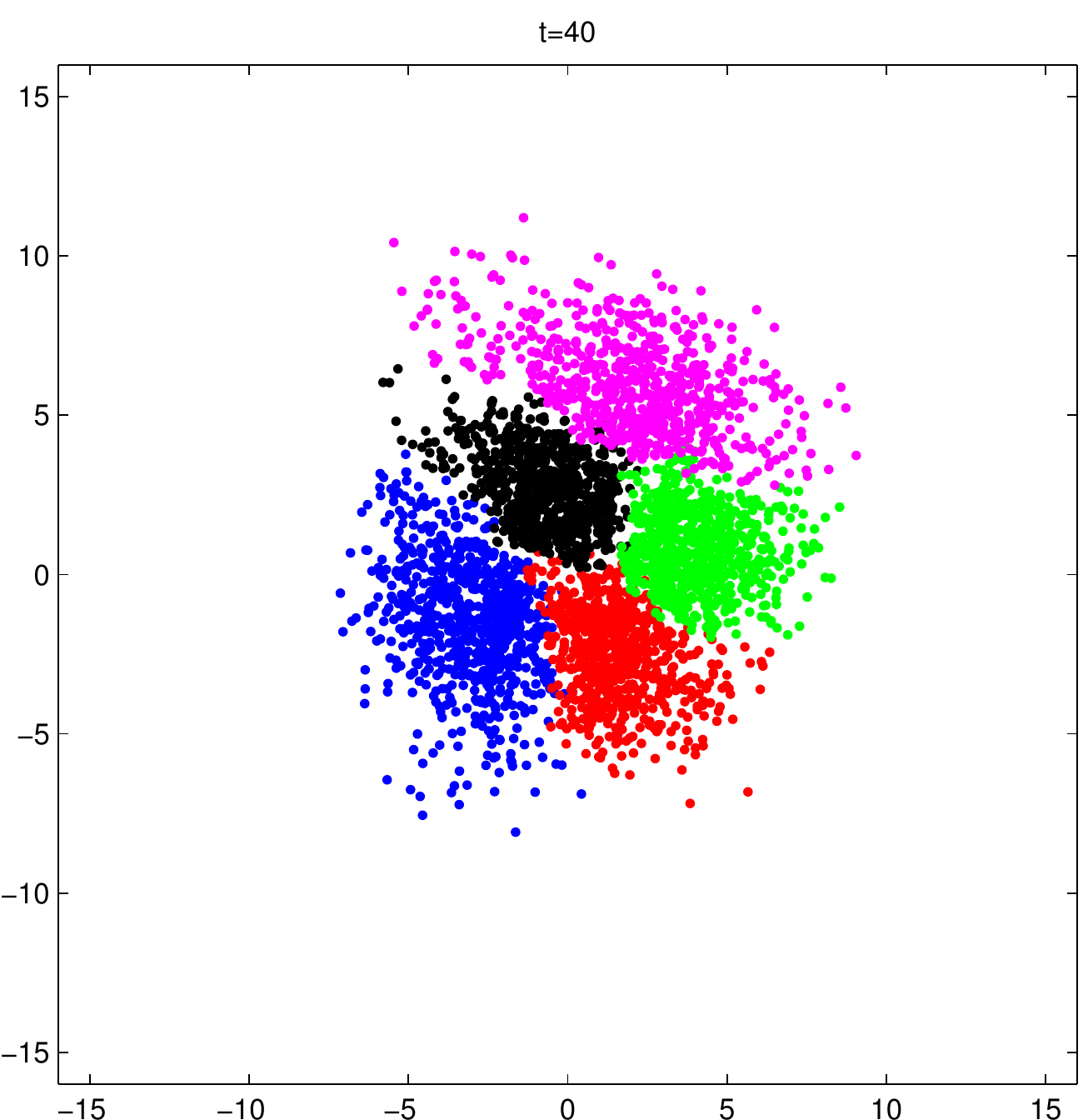}
\includegraphics[width=4cm]{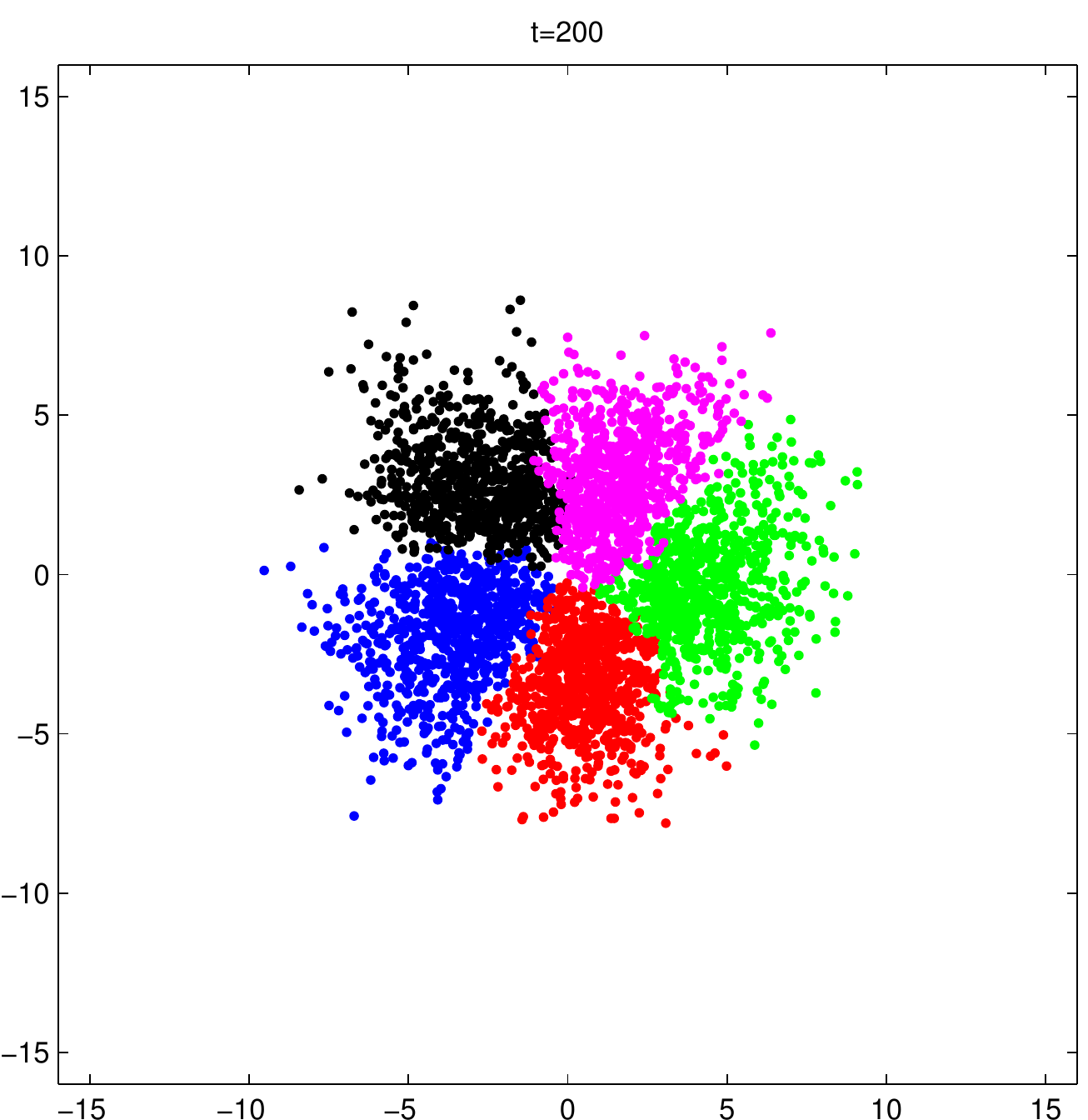}
\end{center}
\caption{ \label{fig:twodexemplar} The distribution of exemplars at four different times in a simulation of the exemplar model in two dimensions with 6 categories. The bottom right plot is representative of the equilibrium distribution of the system. }
\end{figure}

In Figure~\ref{fig:twodexemplar} we show the results of the exemplar simulation. The system was initialized with 50 exemplars at the same location for each category, each of weight $10^{-3}$. The five categories were located at the points $(-5,-5), (0,0), \ldots, (15,15)$.  Exemplars were deleted when their weight decreased below $10^{-4}$.

\begin{figure}[ht]
\begin{center}
\includegraphics[width=4cm]{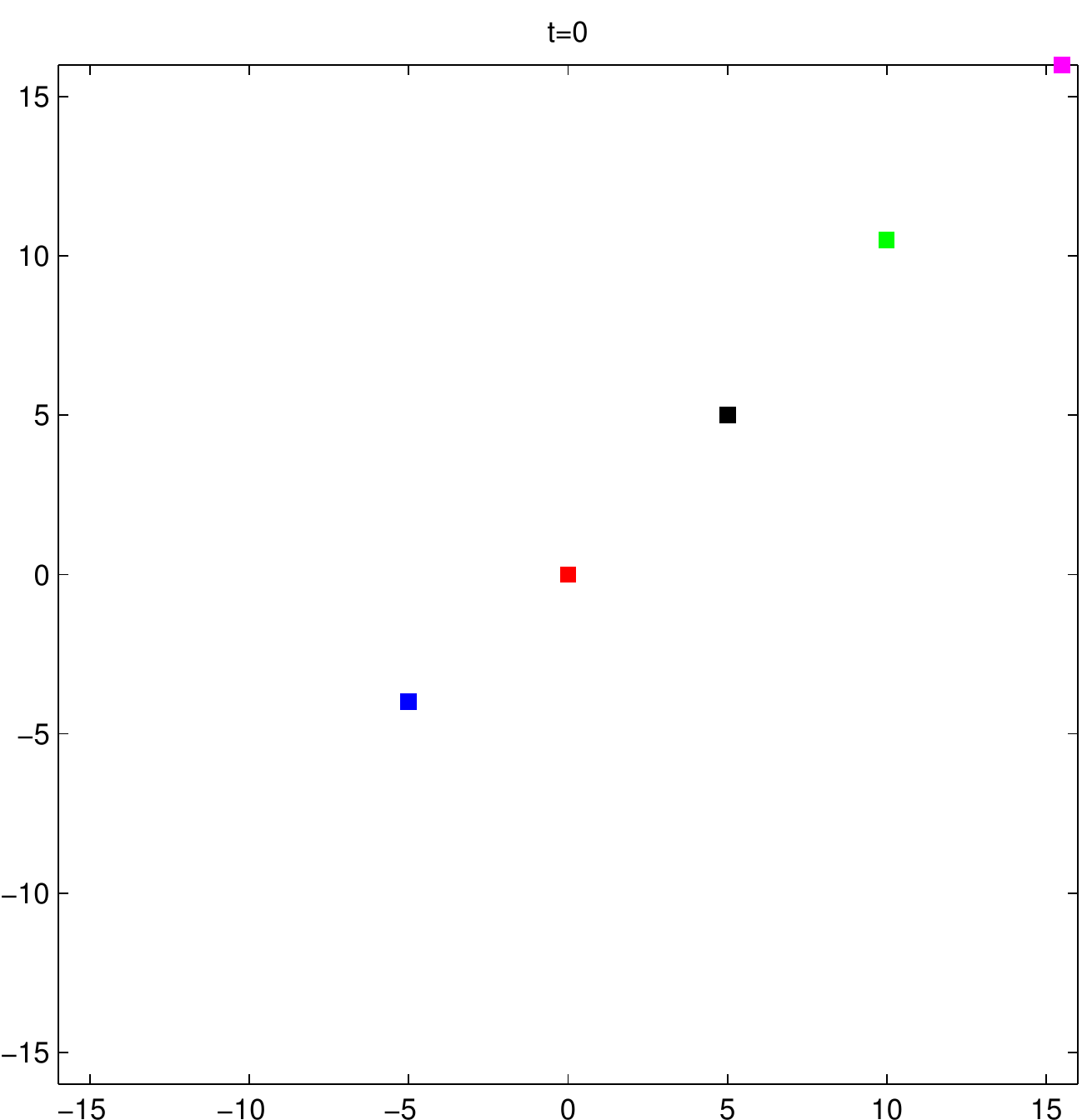}
\includegraphics[width=4cm]{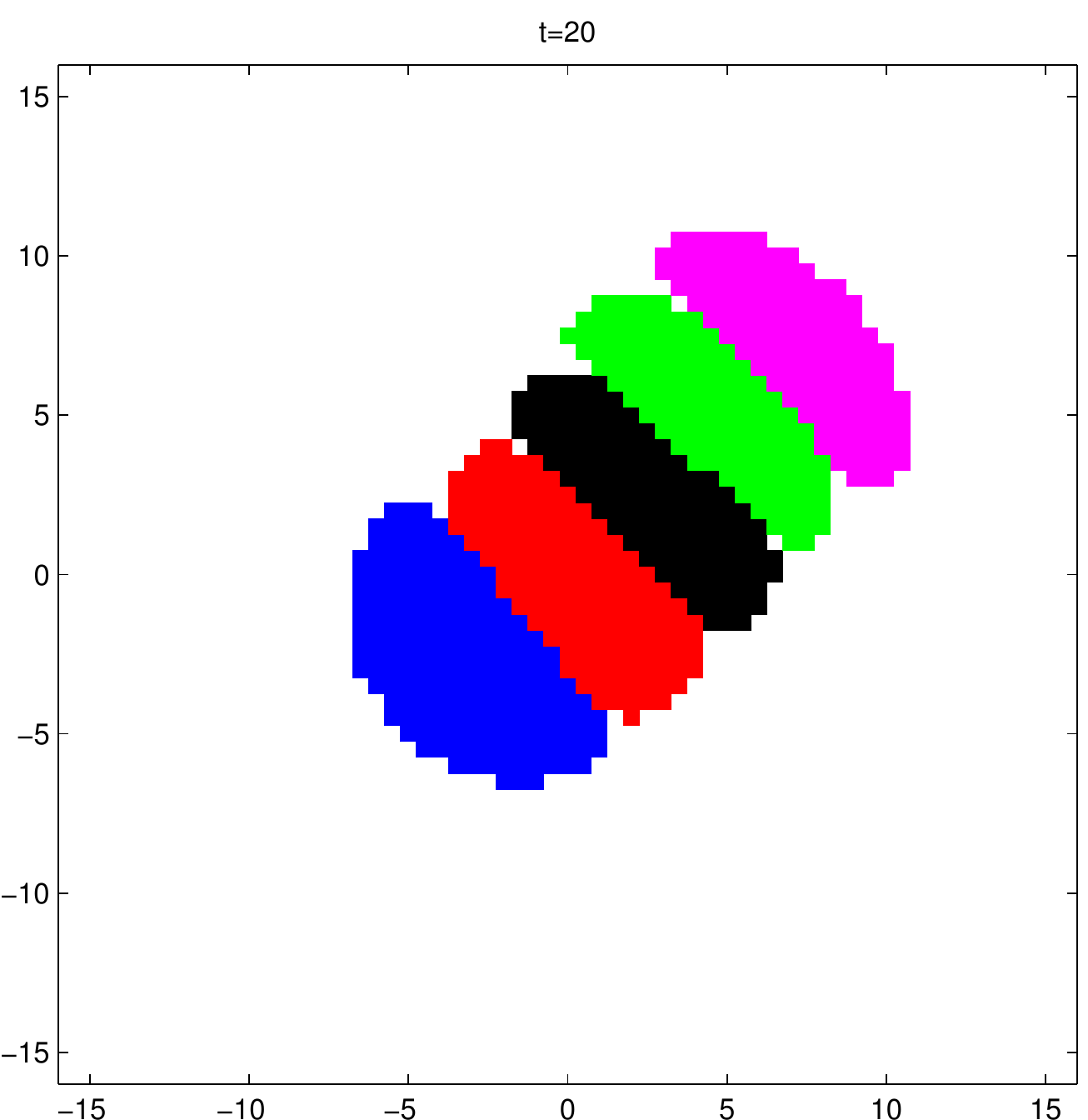}
\includegraphics[width=4cm]{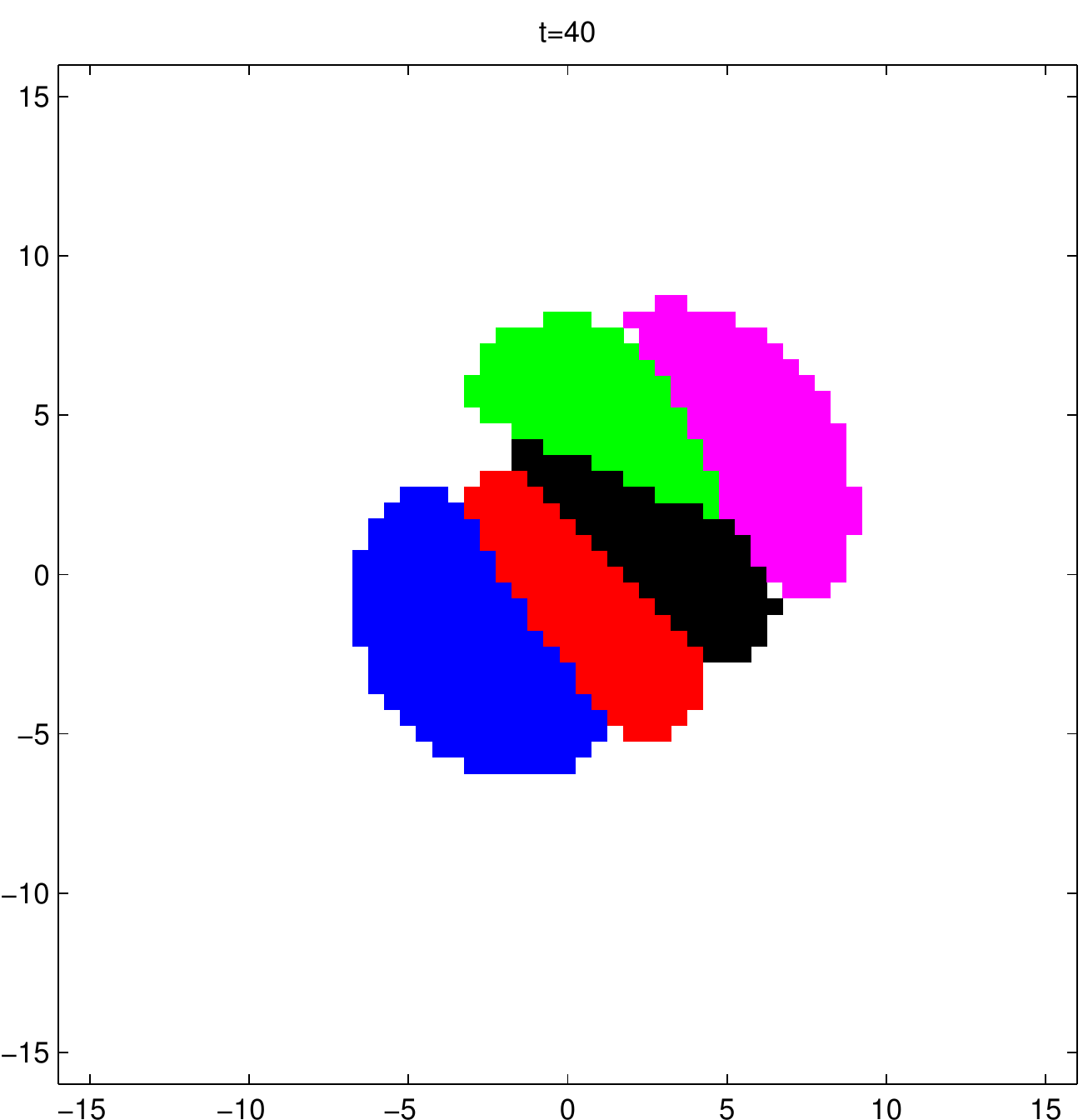}
\includegraphics[width=4cm]{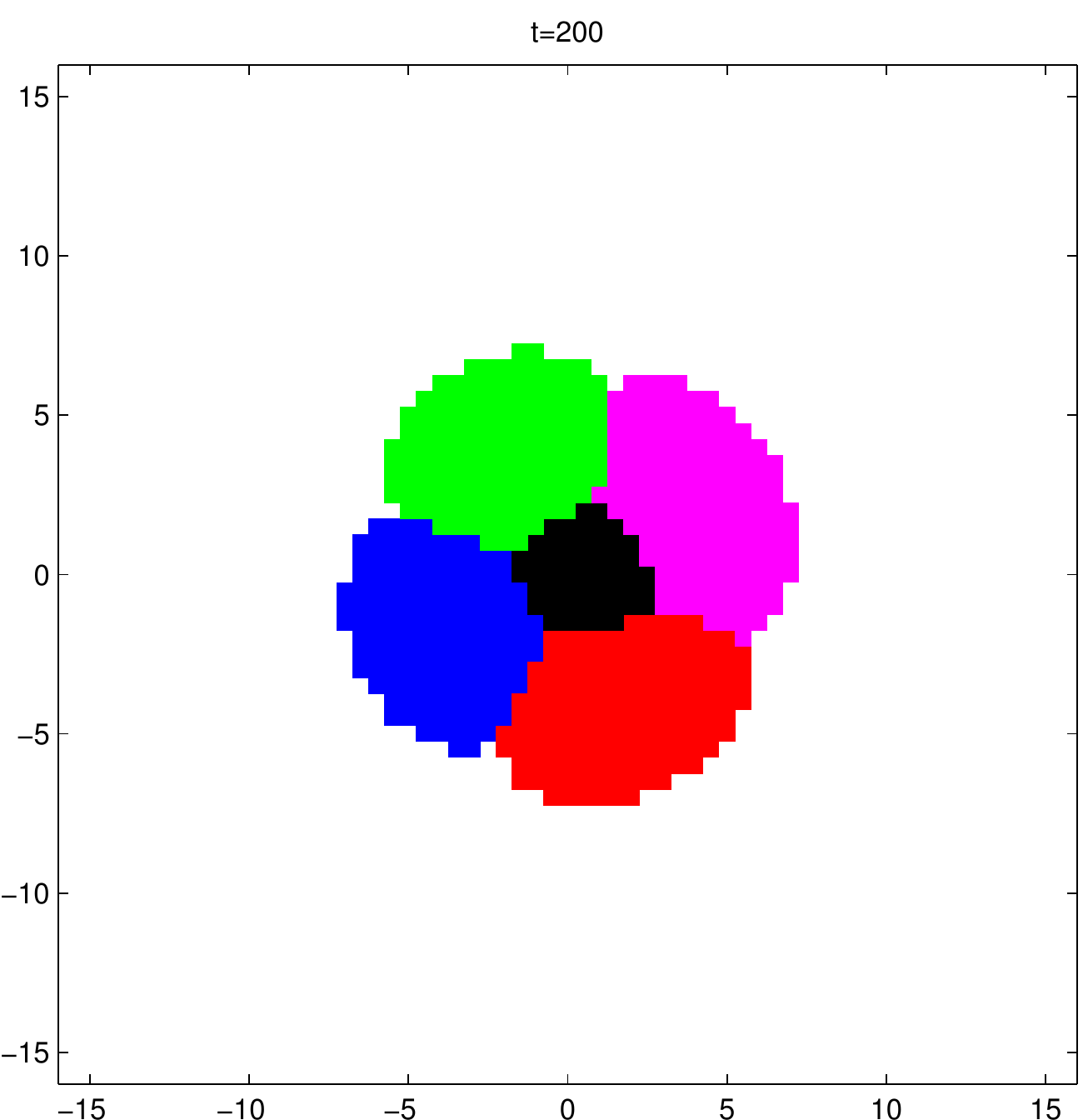}
\caption{ \label{fig:twodfield} Peaks in the exemplar density field at four different times in a simulation of the field model in two dimensions with 6 categories.  The bottom right plot is close to equilibrium.}
\end{center}
\end{figure}

The corresponding simulation for the field model is shown in Figure~\ref{fig:twodfield}. 
%
We observe that both the exemplar model and the field model produce stable distinct categories with the Competition with Discards regimes.

\section{Conclusions and Discussion}

To  conclude, we give an example of the three models we have considered at work within a linguistic context.
 Consider the following example in which the two relevant categories are the English vowel sounds \emph{i} and \emph{e} within the context of the words \emph{bit} and \emph{bet}. A speaker intends to utter the sentence 
\begin{center}
``The dog b{\bf{i}}t me!"
\end{center}
but what is received by the listener is
\begin{center}
``The dog b{\bf{X}}t me!"
\end{center}
where $X$ is some indeterminate vowel sound between an unambiguous \emph{i} and an unambiguous \emph{e}. We will assume that the uttered X is closer to \emph{e} than to \emph{i}.

\textit{ \textbf{ No Competition.}} The listener classifies the vowel sound as \emph{i}, because \emph{bit} is obviously what is intended in this context.
 Thereafter X is stored as an exemplar of \emph{i}.

\textit{ \textbf{Pure Competition.}} Since X is closer to an \emph{e} than an \emph{i}, is classified as an \emph{e}. Thereafter, X is stored as an exemplar of \emph{e}. The meaning of the sentence plays no role in the classification.

\textit{ \textbf{Competition with Discards.}} Since X is closer to an \emph{e} than an \emph{i}, X is tentatively classified as an \emph{e}. However, since the sentence doesn't make any sense with X as \emph{e}, (dogs can't bet people) X is not stored as an exemplar of the category \emph{e}. The sound X is discarded and not stored as an exemplar in any category.
The conclusions from  our modelling are as follows: 

\vspace{-0.2cm}
\begin{itemize} \setlength{\itemsep}{-.1cm}
\item Under the No Competition regime the categories \emph{i} and \emph{e} will merge in this context, in agreement with Blevins and Wedel (2009).
\item Under the Pure Competition Regime the categories will either merge, or together be unstable, (depending on the value of parameter $p$). This is in disagreement with Blevins and Wedel (2009).
\item Under the Competition with Discards regime categories will remain distinct and stably located in phonetic space. 
\end{itemize}
This last point provides support for the idea of exemplar rejection as the mechanism for category stability as described in qualitative terms by Labov (1994) and Silverman (2006).
\vspace{-0.2cm}

\section{Acknowledgments}

The author thanks A. Wedel, J. F. Williams, J. Alderete  and the Vancouver Phonology Group for comments on an earlier draft. The author was supported by a Discovery Grant awarded by NSERC Canada and a Tier II Canada Research Chair.

\nocite{wedel2006,ettlinger,blevins_wedel,labov}

\bibliographystyle{apacite}

\setlength{\bibleftmargin}{.125in}
\setlength{\bibindent}{-\bibleftmargin}


\def\cprime{$'$} \def\cprime{$'$}

\end{document}